\pdfoutput=1

\documentclass[11pt]{article}

\usepackage[final]{acl}

\usepackage{times}
\usepackage{latexsym}
\usepackage{graphicx}
\usepackage{xcolor,colortbl}
\usepackage{booktabs}
\usepackage{amssymb}
\usepackage{multirow}
\usepackage{threeparttable}
\usepackage[T1]{fontenc}

\usepackage[utf8]{inputenc}

\usepackage{microtype}

\usepackage{inconsolata}

\title{Gap-Filling Prompting Enhances Code-Assisted Mathematical Reasoning}

\author{Mohammad Ghiasvand Mohammadkhani\\
Amirkabir University of Technology\\
\texttt{mohammad.ghiasvand@aut.ac.ir}
}

\begin{document}
\maketitle
\begin{abstract}
	Despite the strong performance of large language models (LLMs) in tasks like mathematical reasoning, their practical use is limited by high computational demands and proprietary restrictions. Chain-of-thought (CoT) and program-of-thought (PoT) fine-tuning are common methods to transfer LLM knowledge to small language models (SLMs). However, CoT often leads to calculation errors in SLMs, while PoT has shown more promise. While most PoT-based approaches focus on direct problem-to-code conversion or extracting only the key information from questions and then providing code solution for it, this work emphasizes filling the gaps in the question to clearly illustrate the solution path, which can be challenging for an SLM to understand when such information is not explicitly provided. Therefore, this paper introduces \textit{\textbf{G}ap-\textbf{F}illing \textbf{P}rompting} (GFP), a novel two-step prompting strategy designed to enhance the problem-solving process for SLMs. The first step identifies these gaps and provides hints for filling them, while the second step adds the hints to the question to generate a final code solution. Experimental results on two benchmark datasets demonstrate that GFP significantly improves the mathematical reasoning abilities of SLMs.\footnote{\footnotesize{Code and data released at  \url{https://github.com/mghiasvand1/Gap-Filling-Prompting}}}

\end{abstract}

\section{Introduction}
\begin{figure}[!ht]
	\centering
	\includegraphics[width=0.75	\columnwidth,height=0.6\textheight,keepaspectratio]{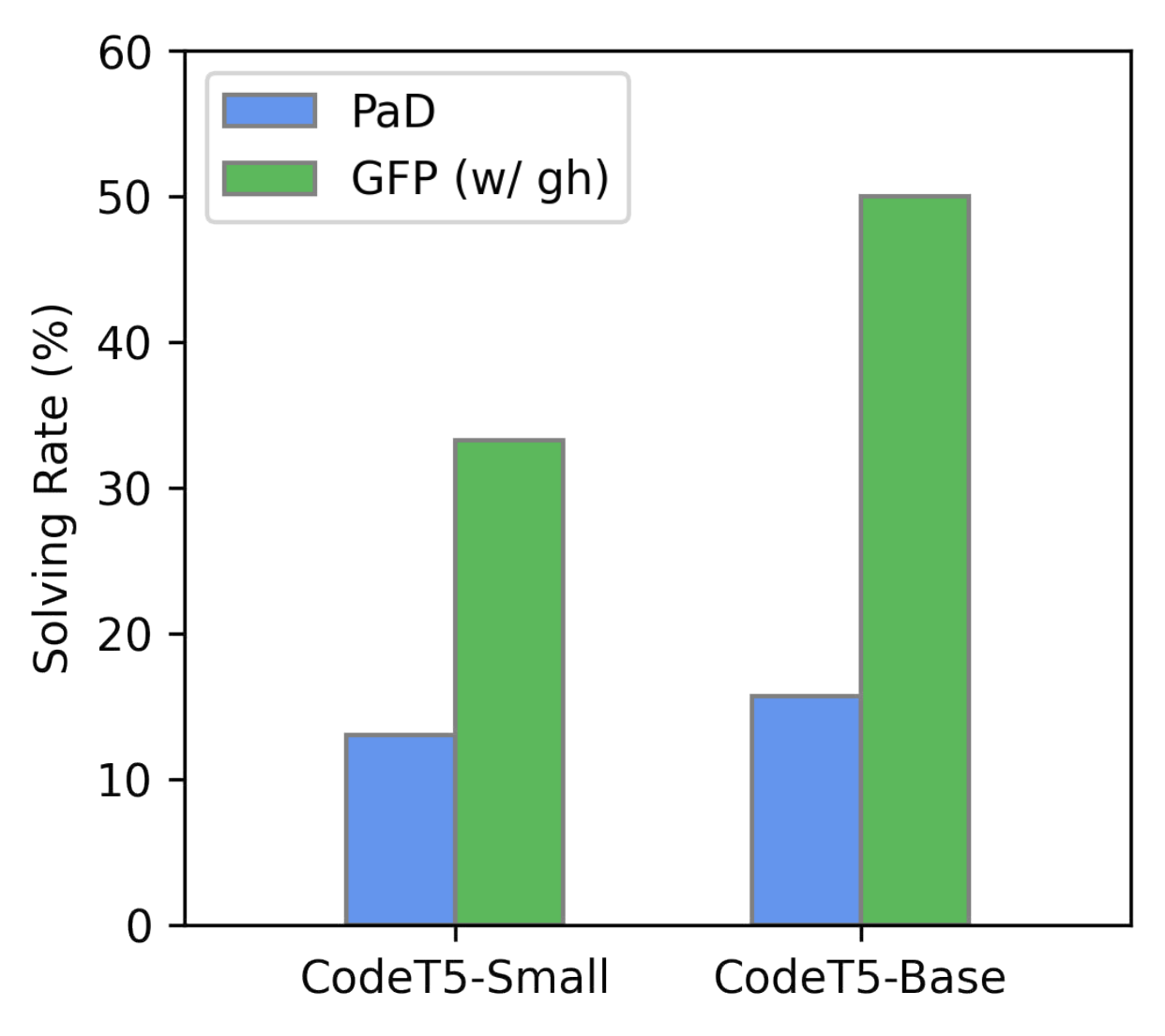}  
	\caption{Experimental results on GSM8K test data show that the code generator models could benefit significantly from gold hints (GFP w/ gh), which fill question gaps, leading to a sharp performance boost over the baseline (PaD), which uses only the original question.}
	\label{fig:emperical}
\end{figure}
LLMs \cite{touvron2023llama, chowdhery2023palm} have excelled in mathematical reasoning tasks, but their large size or proprietary nature creates significant challenges for practical deployment. 
A resource-friendly approach to this issue is distilling the abilities of LLMs into SLMs \cite{ho2023large, magister2023teaching}. CoT and PoT methods are common, where data from LLMs is used to generate step-by-step answers or code, followed by fine-tuning SLMs to enhance reasoning. While CoT fine-tuning often results in high calculation errors \cite{zhu2024key} due to transformers' reliance on next-token prediction, recent research has shifted toward PoT-based methods. In these, the final solution is provided as Python code, which, when executed, returns the answer, minimizing calculation errors. Errors here are mainly of two types: understanding errors and compilation errors, with the former being more prevalent.

Most recent works focused primarily on the direct transformation of problems into code \cite{zhu2024pad, zhu2024distilling}, or at most, extracting key information \cite{zhu2024key} rather than enriching the question with hints that cover its gaps. However, With the development of code-specific language models like CodeT5 \cite{wang2021codet5}, it’s clear that understanding a problem is harder than writing the code to solve it. This work focuses on improving code-assisted mathematical reasoning by addressing gaps in questions, which makes coding easier for SLMs. This paper proposes a novel prompting strategy, GFP, for synthesizing data and fine-tuning SLMs with LLM knowledge. GFP splits reasoning into two stages: first, the SLM generates hints to fill gaps in the question; then, the enhanced question is used to produce Python code, which can return the final result when executed. In summary, this paper makes several key contributions:
\begin{itemize}
    \item It presents insights by empirically studying how gold hints, compared to the original question, can influence the SLM in providing the correct code solution.
    \item This work introduces GFP, a straightforward yet effective code-assisted prompting framework for mathematical reasoning, motivated by the need to fill key gaps in questions in order to provide a proper code solution.
    \item The effectiveness of the GFP framework is shown by experimental results on two benchmark datasets, compared to strong baselines using standard, non-augmented training data.
\end{itemize}

\section{Related Works} 
\subsection{LLMs for Mathematical Reasoning}
The development of step-by-step thinking in LLMs began with \citet{wei2022chain}, who demonstrated that Chain-of-Thought could effectively improve the reasoning performance of LLMs. To avoid calculation errors in Chain-of-Thought, \citet{chen2022program} formalized the reasoning process into a program. Progressive hint prompting \cite{zheng2023progressive} enhances LLM performance by providing approximate hints at each step, refining these hints after each iteration to make them more accurate. Least-to-most prompting \cite{zhou2022least} breaks down complex problems into a series of simpler subproblems, solving them sequentially. Role-play prompting \cite{kong2024better} has also encouraged divergent thinking in LLMs, positively influencing the reasoning process. The hint-before-solving approach \cite{fu2024hint} has demonstrated the benefits of offering hints beforehand to improve problem-solving effectiveness.

\subsection{Reasoning Distillation from LLMs}
\label{sec:rdw}
Reasoning distillation involves first prompting a strong LLM to generate a distillation dataset, which is then used to fine-tune SLMs to improve their reasoning performance. For example, \citet{ho2023large} considered the solution path in the format of CoT step-by-step thinking. \citet{fu2023specializing} provided a thorough analysis of specializing SLMs through CoT reasoning, while \citet{shridhar2023distilling} addressed this task using the decomposer-integrator approach to break down and integrate subproblem answers. \citet{zhu2024pad} proposed Program-aided Distillation (PaD) as a strong baseline that combines code-assisted mathematical reasoning with high-rate data augmentation and incorporates additional features such as a verifier. Similarly, \citet{zhu2024distilling} applied a new approach named Equation-of-Thought to address and further process the data encountered during compilation errors in PoT. Additionally, \citet{zhu2024key} employed a three-step strategy directly inspired by \citet{zhong2024achieving}, which involves first extracting the core question, then identifying and extracting the key information, and finally providing the Python code. However, most of these works focus on a direct problem-to-code approach, applying effective data augmentation, or only extracting key information. The importance of understanding question gaps for SLMs is often neglected, despite the fact that these gaps can be filled by hints to improve the model's understanding in producing the final code solution.

\section{Methodology}
\begin{table*}[!ht]
	\centering
	\small
	\begin{tabular}{l|c|cc|c} 
		\toprule 
		\textbf{Models} & \textbf{\#Params} & \textbf{GSM8K} & \textbf{MultiArith} & \textbf{AVG}\\
		\midrule 
		\rowcolor[rgb]{0.93,0.93,0.93}
		\multicolumn{5}{l}{\textit{Pre-Trained LLMs}} \\
		\texttt{gpt-4-0613} & ? & 92.0 &  97.8 & 94.9\\
		LLaMA-2 ~\citep{touvron2023llama} & 70B & 56.8 & 90.2 & 73.5\\
		CodeLLaMA~\citep{roziere2023code} & 7B & 34.0 & - & 34.0\\
		PaLM~\citep{chowdhery2023palm} & 60B & 29.9 & 75.0 & 52.45\\
		Platypus-2~\citep{lee2023platypus} & 7B & 14.4 & - & 14.4\\
		\midrule 
		\rowcolor[rgb]{0.93,0.93,0.93}
		\multicolumn{5}{l}{\textit{Fine-tuned SLMs}} \\
		\citet{ho2023large} & 0.3B & 3.11 & 6.11 & 4.61\\
		\citet{fu2023specializing} & 0.25B & 13.4 & 29.7 & 21.55\\
		\citet{fu2023specializing} & 0.76B & 20.2 & 38.5 & 29.35\\
		\citet{shridhar2023distilling} & 0.77B & 17.89 & - & 17.89\\
		PaD \cite{zhu2024pad} & 0.06B & 13.0 & 26.5 & 19.75\\
		PaD \cite{zhu2024pad} & 0.22B & 15.7 & 25.0 & 20.35\\
		PaD \cite{zhu2024pad} & 0.77B & 21.7 & 34.3 & 28.0\\
		\midrule
		GFP (Ours) & 0.06B & \textbf{20.07} & \textbf{69.0} & \textbf{44.53}\\
		GFP (Ours) & 0.22B & \textbf{24.86} & \textbf{73.0} & \textbf{48.93}\\
		\midrule 
	\end{tabular}
	\caption{Overall test set performance.}
	\label{tab:main_results}
\end{table*}
This section provides an overview of the operational procedures of the GFP framework. The primary goal of the mathematical reasoning task is to take a math problem as input and return a number as the final answer. GFP achieves this by first synthesizing data using the teacher LLM, which generates hints and code for each data point. The student SLMs are then fine-tuned to fill in question gaps by generating hints, followed by providing the code solution and executing it to obtain the final numeric result.

\subsection{Training}  
Since obtaining both hints and code solutions for each data point in the training data is required based on the GFP framework architecture, the teacher LLM is prompted using the template described in Appendix \ref{app:prompt} to generate hints and Python code for each training data point. For each data point, it is important to verify that the Python code returned by the LLM produces the correct numerical result by executing the code and checking the value of the \textit{“result”} variable. If the correct answer is not obtained, the data point is completely removed from the training data.\\  
Next, two different SLMs are employed to distill the teacher LLM’s capabilities into them. This process involves designing distinct targets and inputs for each step and initializing parameter \( \theta \). Each step model is fine-tuned separately to minimize the following Cross-Entropy loss function during training.  
\begin{equation}
	\mathcal{L}(x,y) = -\sum_{t=1}^{n} \log p_{\theta}(y_{t} | x, y_{<t})
\end{equation}
where $n$ represents the length of the target sequence $y$, and $y_{<t}$ denotes the tokens generated previously. For the first SLM, the inputs are the original questions, and the target outputs are the hints returned by the teacher LLM for that problem, with the hints separated by \textit{“ \& ”}. \\
To form the input for the second SLM, which is responsible for code generation, the input consists of the concatenation of the original question, followed by \textit{“ \#\# ”}, and then the gap-filling hints for that question, with the hints also separated by \textit{“ \& ”}. The target output for this SLM is the code.

\subsection{Inference}
In the inference stage, for each item in the test set, the original question is first fed into the hint generator model to obtain the corresponding hints. The input for the code generator model is then constructed, similar to the process in the training stage, to form a rich text that fills in the question gaps. Afterwards, the Python code is run through an interpreter, and the stored value in the \textit{“result”} variable is returned as the final answer.

\section{Experiments and Results}

\subsection{Experimental Setup}
In this work, the GPT-4 model from the official OpenAI API was used as the teacher LLM for data synthesis. The framework employed Flan-T5 \cite{chung2024scaling}, an instruction-tuned version of the original T5 model \cite{raffel2020exploring}, as the hint generator for filling question gaps in the first phase, and CodeT5 for code generation in the second phase; both were sourced from the Hugging Face Transformers library\footnote{\footnotesize{\url{https://github.com/huggingface/transformers}}}. The teacher LLM temperature was set to 0 to ensure reproducibility, and its response format was configured to JSON. All SLMs were fine-tuned with a batch size of 8 over 10 epochs using a learning rate of 3e-4, utilizing the Kaggle\footnote{\footnotesize{\url{https://kaggle.com/}}} NVIDIA Tesla T4 16GB GPU.

\subsection{Datasets}
For training, only the training split of GSM8K \cite{cobbe2021training}, containing 7,470 data points, was used without any data augmentation. For each data point in the dataset, hints and code solutions were constructed, and data points that resulted in compilation errors or incorrect answers from the generated Python code by the teacher LLM were removed. This process reduced the final number of data points to 7,254. GFP was evaluated on the test splits of the GSM8K and MultiArith \cite{roy2015solving} benchmarks, with 1,315 and 600 test data points, respectively.

\subsection{Baselines}
The baselines are divided into two groups: pretrained LLMs and smaller, distilled models from prior research. For the LLMs, comparisons were made against \texttt{gpt-4-0613}, LLaMA-2 \cite{touvron2023llama}, CodeLLaMA \cite{roziere2023code}, PaLM \cite{chowdhery2023palm}, and Platypus-2 \cite{lee2023platypus}. For the smaller finetuned models, since the framework does not use data augmentation and relies on standard datasets, only baselines from works without data augmentation were included to ensure a fair comparison. Therefore, comparisons were made to the finetuned SLMs from \citet{ho2023large}, \citet{fu2023specializing}, \citet{shridhar2023distilling}, and \citet{zhu2024key}, with their methods detailed in Section \ref{sec:rdw}.

\subsection{Results}
The paper results and comparisons with other methods are available in Figure \ref{fig:emperical} and Table \ref{tab:main_results}. The figure pertains to the first contribution, which is an empirical study on the motivation for addressing question gaps. It demonstrates that as more correct hints are provided, the final code accuracy improves, making it easier for the code generator model to generate correct code. In the implementation, it is notable that only the base \textit{(0.25B)} version of Flan-T5 serves as the hint generator across all cases. Due to the critical nature of this stage, reducing the model size significantly impacts performance. Experiments were conducted using both the \textit{small (0.06B)} and \textit{base (0.22B)} versions of CodeT5 as the code generator model. The framework size shown in the table corresponds to the second-step model (CodeT5). As indicated by the table, the GFP framework significantly outperforms the baselines for both \textit{small} and \textit{base} sizes of the code generator model. The \textit{base} framework surpasses PaD’s \textit{base} and \textit{large (0.77B)} versions by $9.16\%$ and $3.16\%$ in GSM8K, $48.0\%$ and $38.7\%$ in MultiArith, and by $28.58\%$ and $20.93\%$ on average, despite the notable difference in parameter size between the \textit{base} and \textit{large} versions. Additionally, the model achieves competitive results against larger pretrained LLMs despite the substantial difference in parameter size and efficiency. Even when the \textit{small} CodeT5 model is used in the framework, performance does not drop drastically and still surpasses the baselines, underscoring the effectiveness of gap-filler hints across different model sizes.

\section{Conclusion}
This paper introduces GFP, an effective yet straightforward framework for mathematical reasoning. An LLM was utilized as the teacher for data synthesis, followed by fine-tuning SLMs to first generate hints to fill question gaps and then provide code solutions. Despite the absence of data augmentation—unlike previous works—and the simplicity of the framework's architecture, GFP outperforms strong baselines in mathematical reasoning with SLMs. This method sets a new benchmark, highlighting promising avenues for future research in this field.

\section{Limitations}
Although this straightforward framework demonstrates strong performance, it has limitations that suggest areas for future research. The approach is promising for multi-step reasoning with gaps but may struggle with simpler datasets that feature single-stage solutions or minimal gaps, where hints are largely unnecessary as most solution paths can be inferred directly from the questions. This limitation arises from confusion in the hint generator model, indicating the need for flexibility based on input problem types. Additionally, since the method was applied only to standard training data, implementing a data augmentation strategy similar to previous baselines could improve overall performance. While this work focuses on distilling mathematical reasoning to SLMs, the potential to adapt this approach to enhance LLM performance in complex reasoning tasks also appears promising.

\bibliography{custom}

\appendix

\section{Prompt Template}
\label{app:prompt}
\begin{figure*}
	\centering
	\includegraphics[width=6in]{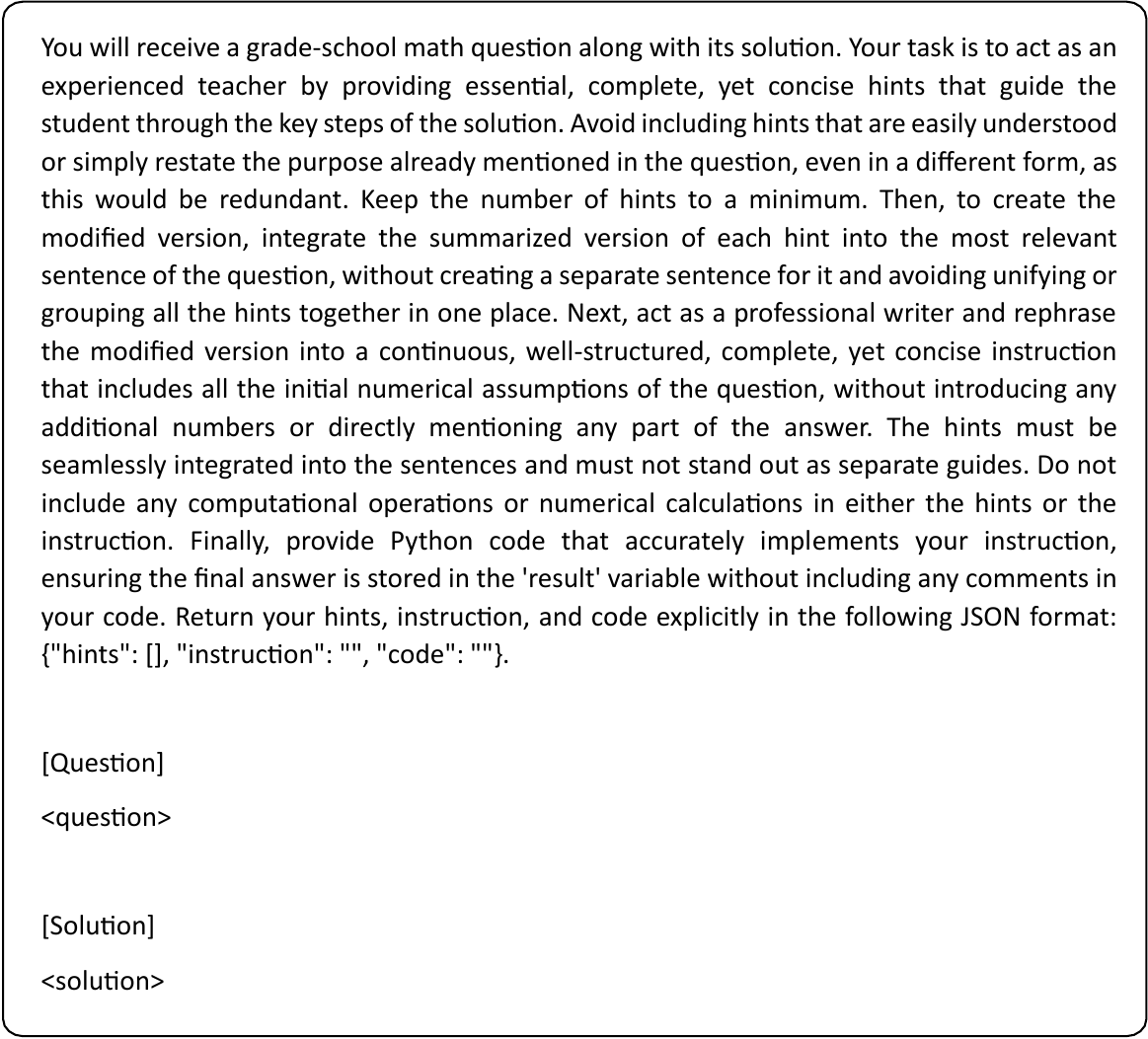}
	\caption{The Prompt Template for Data Synthesis}
	\label{fig:fig2}
	
\end{figure*}

Figure \ref{fig:fig2} shows the prompt template for generating synthetic data from the teacher LLM, where the user message is included, and the system message is left empty. The tags \textit{"<question>"} and \textit{"<solution>"} are replaced by the original question and its explained solution, respectively.

\end{document}